\documentclass[sigconf, nonacm]{acmart}

\AtBeginDocument{%
  \providecommand\BibTeX{{%
    \normalfont B\kern-0.5em{\scshape i\kern-0.25em b}\kern-0.8em\TeX}}}

\usepackage{array}
\newcolumntype{Y}{>{\centering\arraybackslash}X}
\usepackage{booktabs} 
\usepackage{verbatim}

\usepackage{amsfonts}
\usepackage{graphicx}

\usepackage{cases}
\usepackage{bm}
\usepackage{multirow}
\usepackage{subfig}
\usepackage{tabularx}
\usepackage{balance}

\usepackage{amsmath}

\usepackage{url}

\usepackage{enumitem}

\begin{document}

\title{SIM-Trans: Structure Information Modeling Transformer for Fine-grained Visual Categorization}


\author{Hongbo Sun}
\affiliation{%
  \institution{Wangxuan Institute of Computer Technology, Peking University}
  \city{Beijing}
  \country{China}}
\email{sunhongbo@pku.edu.cn}

\author{Xiangteng He}
\affiliation{%
  \institution{Wangxuan Institute of Computer Technology, Peking University}
  \city{Beijing}
  \country{China}}
\email{hexiangteng@pku.edu.cn}

\author{Yuxin Peng}
\authornote{Corresponding author.}
\affiliation{%
  \institution{Wangxuan Institute of Computer Technology, Peking University}
  \city{Beijing}
  \country{China}}
\email{pengyuxin@pku.edu.cn}

\renewcommand{\shortauthors}{Hongbo Sun, Xiangteng He, \& Yuxin Peng}
\pagestyle{plain}



\begin{abstract}

Fine-grained visual categorization (FGVC) aims at recognizing objects from similar subordinate categories, which is challenging and practical for human’s accurate automatic recognition needs. Most FGVC approaches focus on the attention mechanism research for discriminative regions mining while neglecting their interdependencies and composed holistic object structure, which are essential for model’s discriminative information localization and understanding ability. To address the above limitations, we propose the \textbf{\emph{Structure Information Modeling Transformer (SIM-Trans)}} to incorporate object structure information into transformer for enhancing discriminative representation learning to contain both the appearance information and structure information. Specifically, we encode the image into a sequence of patch tokens and build a strong vision transformer framework with two well-designed modules: (i) the \textbf{\emph{structure information learning (SIL) module}} is proposed to mine the spatial context relation of significant patches within the object extent with the help of the transformer’s self-attention weights, which is further injected into the model for importing structure information; (ii) the \textbf{\emph{multi-level feature boosting (MFB) module}} is introduced to exploit the complementary of multi-level features and contrastive learning among classes to enhance feature robustness for accurate recognition. The proposed two modules are light-weighted and can be plugged into any transformer network and trained end-to-end easily, which only depends on the attention weights that come with the vision transformer itself. Extensive experiments and analyses demonstrate that the proposed SIM-Trans achieves state-of-the-art performance on fine-grained visual categorization benchmarks. The code is available at \url{https://github.com/PKU-ICST-MIPL/SIM-Trans_ACMMM2022}.

\end{abstract}

%




\maketitle

\section{Introduction}

Fine-grained visual categorization (FGVC) task \cite{wei2021fine} targets at recognizing object into specific subcategory from a given basic category, such as identifying bird species \cite{cub200}. It is different from generic image classification which only needs to predict the basic category, such as “Bird”. There exist much more challenges due to its intrinsic large intra-class variance and small inter-class variance. For example, “Artic Tern” and “Common Tern” belong to the “Bird” basic category and only have subtle difference in the tail and beak, which are hard to distinguish. Thus, locating discriminative regions to extract features and designing high-order robust features are researched for addressing the above problem. The localization based methods are widely studied for its better interpretation and promising performance.

Early works \cite{berg2013poof, zhang2014part, branson2014bird} are designed to localize discriminative regions with the help of human annotations, i.e., bounding box of object or part annotations. Whereas, human annotations for fine-grained image classification are hard to obtain due to the strict requirements for expertise. Aiming at addressing this problem, a lot of researches \cite{zheng2017learning, peng2018object, he2018fast, yang2018learning, wang2018learning, he2019and, chen2019destruction, ge2019weakly, ding2019selective, sun2020fine, zhou2020look,  du2020fine, wang2021dynamic} have been conducted on the weakly-supervised fine-grained visual categorization task, which only utilizes the image category labels. They mainly detect semantic parts explicitly \cite{zheng2017learning, yang2018learning, he2018fast, ge2019weakly} or conduct saliency regions positioning implicitly \cite{peng2018object, chen2019destruction, ding2019selective, zhou2020look}, which follows the feature extraction and fusion for final classification. The first class of methods such as \cite{he2018fast, ge2019weakly} mainly adopt the region proposal network to obtain the location of discriminative image regions. The selected image regions are resized into fixed size and input into the backbone network for feature extraction and classification. And the second class of methods such as \cite{peng2018object, ding2019selective} exploit the attention mechanism for salient regions detection and utilization, which can be flexibly designed along with the backbone network. However, the above methods generally ignore the relation among discriminative regions within object in the model designing. It may cause bad localization results with large area of irrelevant background, which leads to a sharp drop of classification performance. Meanwhile, when reviewing the intrinsic structure of the CNN based FGVC methods, we find that the stacked convolution and pooling operations bring both the expansion of the receptive field and the degradation of spatial discrimination. Large continuous image areas are focused on and discriminative details are generally overlooked, which are essential for distinguishing subtle difference in fine-grained visual categorization. 

Recently, vision transformer (ViT) ~\cite{dosovitskiy2020image} and its variants present a new solution of encoding image into a sequence of patch tokens for recognition, which has achieved promising performance. The multi-head self-attention mechanism in transformer provides long-range dependency to enhance the interaction among image patches. And the discriminative patch information is kept with the deeper of the transformer layer for deciding the final classification. Thus, the vision transformer can alleviate the aforementioned problems in CNN based FGVC methods to some extent. RAMS-Trans \cite{hu2021rams} proposes the dynamic patch proposal module to guide region amplification for multi-scale learning in fine-grained visual categorization. However, the above methods mainly focus on significant patch tokens selection while ignoring their relation in the holistic object structure, which is also essential for identifying discriminative regions. For example, models can be puzzled due to lacking the cognitive ability for object structure in many cases ~\cite{zhou2020look}, such as localizing legs of a bird among twigs.

Therefore, we propose the structure information modeling transformer dubbed SIM-Trans to introduce structure information into the vision transformer for fine-grained visual categorization. SIM-Trans attempts to model the context information among regions within the object to highlight discriminative regions for accurate recognition. Firstly, we construct a vision transformer backbone to encode the image into a sequence of patch tokens for feature extraction. Secondly, in order to model the object structure information, we propose the \textbf{\emph{structure information learning (SIL) module}} to mine spatial context relations among discriminative patches within the object extent. Benefiting from the self-attention characteristics of transformer, the attention weight between the cls token (standing for the whole image) and patch token is highly correlated with whether the patch token contains the object information. Thus, discriminative patches can be selected expediently in this way. The relative position relation and semantic relation among patches are calculated to construct the graph of depicting the object structure information, which is further extracted and injected into the backbone by graph convolution. The SIL module can boost the model to learn the object structural composition and highlight significant regions through end-to-end training. Thirdly, to further enhance the feature robustness and discrimination, a \textbf{\emph{multi-level feature boosting (MFB) module}} is designed. We propose to concatenate the features from the last three transformer layers to take advantage of their complementary, which have been injected the object structure information by the aforementioned SIL module. Besides, contrastive learning is introduced to further boost model’s performance, which enhances the feature representation similarity of samples from the same category and weaken that from different categories.

The main contributions made in this paper can be summarized as follows:
\begin{itemize}[leftmargin=30pt]

\item {We propose the structure information modeling transformer dubbed SIM-Trans with two well-designed modules for boosting fine-grained representation learning to contain both the appearance information and structure information.
}

\item {The structure information learning (SIL) module is proposed to mine the spatial context relations among discriminative regions within object extent, which boosts the model’s understanding ability for object structure. The multi-level feature boosting (MFB) module is designed to exploit the complementary of multi-level features and contrastive learning for robust feature representation.}

\item {The proposed SIL and MFB modules are light-weighted, which only depend on the attention weighs that come with the vision transformer itself. They can be easily plugged into any vision transformer backbones and trained end-to-end. Extensive experiments and analyses on two typical fine-grained visual categorization benchmarks demonstrate that our proposed method achieves new state-of-the-art.}

\end{itemize}

The rest of the paper is organized as follows: Section 2 briefly reviews the related work on fine-grained visual categorization and vision transformer. Section 3 elaborates our SIM-Trans approach and Section 4 introduces the experimental results and analyses, as well as ablation studies. Finally, Section 5 concludes this paper.

\section{Related Work}

In this section, we briefly introduce the related work of fine-grained visual categorization and vision transformer.

\begin{figure*}[!th]
  \centering
  \includegraphics[width=\linewidth]{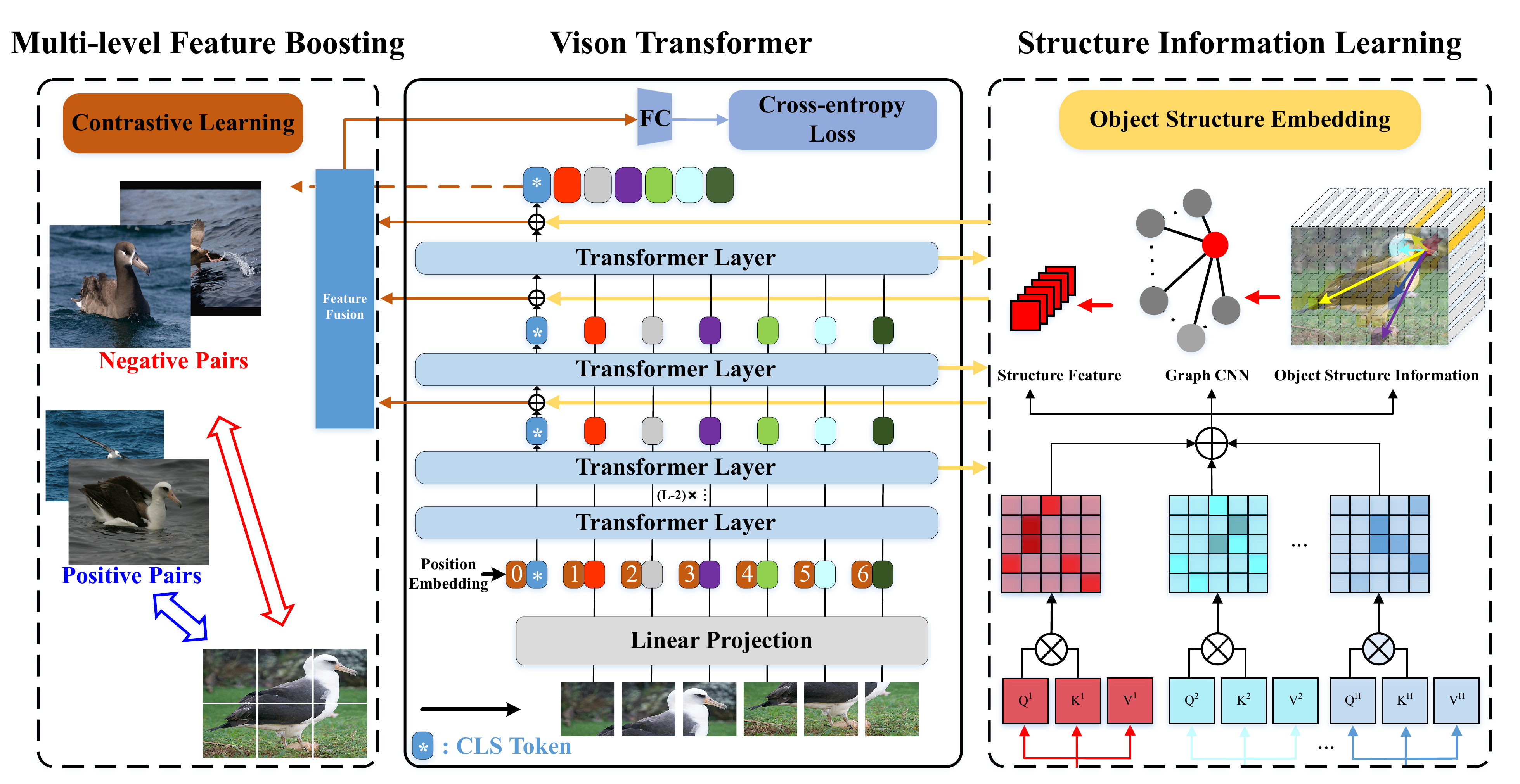}
  \caption{The overall framework of our proposed SIM-Trans.}
  \label{framework}
\end{figure*}

\subsection{Fine-grained Visual Categorization}

Recent approaches mainly focus on the discriminative regions discovery and feature extraction for fine-grained visual categorization \cite{he2018fast, he2019and, ding2019selective, wang2020weakly, wang2020graph, song2020bi}. Ding et al. \cite{ding2019selective} propose S3N to utilize sparse attention to estimate informative regions and extract discriminative and complementary features for classification. He et al. \cite{he2019and} propose M2DRL to automatically determine the location and number of discriminative regions with reinforcement learning paradigm. Song et al. \cite{song2020bi} propose Bi-Modal PMA to capture discriminative parts stage-by-stage with the progressive mask attention model. Rao et al. \cite{rao2021counterfactual} propose a counterfactual attention learning method to obtain more useful attention for fine-grained visual categorization. However, the above methods generally ignore object’s holistic structure information, which is greatly helpful for localizing the whole object extent. In our proposed method, the object structure information is introduced explicitly for highlighting the discriminative regions within object.

\subsection{Vision Transformer}

Recently, many vision transformer methods have been proposed for computer vision tasks. ViT ~\cite{dosovitskiy2020image} is the first work to introduce the pure transformer into image classification, which splits the image into a sequence of patch tokens as input of transformer and achieves promising performance. Zheng et al. \cite{zheng2021rethinking} propose SETR to utilize ViT as the encoder for image segmentation. He et al. \cite{he2021transreid} propose TransReID to introduce the jigsaw patch module and side information embeddings into transformer for object re-identification. Recently, RAMS-Trans \cite{hu2021rams} is proposed for fine-grained visual categorization, which designs the recurrent attention multi-scale transformer to find and amplify significant regions for learning multi-scale features. However, the above vision transformer works generally ignore the spatial relation of patches, which is important for significant patches discovery. Our proposed SIM-Trans incorporates the object structure information into transformer to enhance discriminative patch features for accurate fine-grained visual categorization.

\section{Approach}

Our proposed SIM-Trans is based on the vision transformer with several critical improvements for specializing in fine-grained visual categorization task. The overall framework of SIM-Trans is shown in Figure \ref{framework}. The vision transformer backbone (the middle part) takes image patch tokens and works as an extractor. The self-attention weights of the transformer layer are borrowed by the structure information learning (SIL) module (the right part) to mine the spatial context information of discriminative patches within the object. The multi-level feature boosting (MFB) module (the left part) fuses the feature from different levels to exploit their complementary, and the feature is also boosted by contrastive learning simultaneously.

\subsection{Vision Transformer Backbone} \label{ViT}

The vision transformer backbone is composed of feature extractor and classifier head, which is shown in the middle part of Figure \ref{framework}. Before being input to the feature extractor, the image $x \in {R^{H \times W \times 3}}$, where $H$ and $W$ denotes its height and width respectively, is generally split into $N$ patch tokens through non-overlapping splitting, denoted as $\{ x_{patch}^1,x_{patch}^2,...,x_{patch}^N\}$. However, this splitting way may cause the incomplete neighboring information within the patch due to the hard separation. Thus, we adopt the sliding window splitting method following \cite{he2021transreid}. Specifically, the patch size and the window’s sliding step are denoted as $P$ and $S$, the patch sequence length $N$ for the image $x$ can be calculated as follows:
\begin{equation}
N = {N_H} \times {N_W} = \left\lfloor {\frac{{H - P}}{S} + 1} \right\rfloor  \times \left\lfloor {\frac{{W - P}}{S} + 1} \right\rfloor
\label{slide}
\end{equation}
where $\left\lfloor  \cdot  \right\rfloor$ denotes the floor operation, ${N_H}$ and ${N_W}$ denote the patch numbers in vertical and horizontal directions after splitting. In this way, the local neighboring information preservation problem is alleviated to some degree. 

The patch token $\{ x_{patch}^i\} $ is then projected into $D$-dimensional embedding through linear mapping $F( \cdot )$. For representing the whole image, a learnable $cls$ embedding token ${x_{cls}}$ is introduced and put at the beginning of the input sequence embeddings. To incorporate position information, learnable position embeddings ${E_p} \in {R^{(N + 1) \times D}}$ are added to the input sequence embeddings to get $z_0$ as the input of the first transformer layer as follows:
\begin{equation}
{z_0} = [{x_{cls}},F(x_{patch}^1),F(x_{patch}^2),...,F(x_{patch}^N)] + {E_p}
\end{equation}
The transformer feature extractor is composed of $L$ transformer layers, each of which consists of a multi-head self-attention (MSA) module and a feed forward neural network of two fully connected layers. The output of the ${k_{th}}$ transformer layer is calculated as follows:
\begin{equation}
z_k^{'} = LN(MSA({z_{k - 1}}) + {z_{k - 1}})
\end{equation}
\begin{equation}
{z_k} = LN(FFN(z_k^{'}) + z_k^{'})
\end{equation}
where $LN( \cdot )$ denote the layer normalization \cite{ba2016layer}. The output $cls$ tokens of the last three transformer layers are concatenated and regarded as image representation (more details in Sec. \ref{MFB}), which is forwarded into a classifier head to get the prediction vector $pred(I)$ for the input image $I$. The classification loss is as follows:
\begin{equation}
{L_{CE}} =  - \sum\limits_{I \in S(I)} {(y \cdot \log (pred(I)))}
\end{equation}
where $S(I)$ is the training set and $y$ is the one-hot label of image $I$.

\subsection{Structure Information Learning}\label{SIL}

The vision transformer backbone can achieve promising image classification results, which utilizes the self-attention mechanism to own a global receptive filed. However, the vision transformer framework generally ignores the spatial relation of patches, which is important for identifying discriminative patches in fine-grained visual categorization task. Thus, we propose the structure information learning (SIL) module to incorporate the object spatial context information into the vision transformer and the whole procedure is shown in the right part of Figure \ref{framework}.

Localizing the object extent is the precondition for structure learning. In the transformer layer, the attention weight between the patch token and cls token depicts its importance for the final classification, which is highly correlated with whether the patch token contains the object information. Thus, object can be naturally localized with the help of the attention weights. Suppose the transformer layer has $H$ heads, $Q$ and $K$ are $D$-dimensional query vectors and key vectors of all tokens, then the attention weights can be calculated as follows:
\begin{equation}
At{t_h} = soft\max (\frac{{Q{K^T}}}{{\sqrt {D/H} }})
\end{equation}
where $At{t_h} \in {R^{(N + 1) \times (N + 1)}}$, $h = 1,2,...,H$ and $N = {N_H} \times {N_w}$ is the number of patches. The attention weight between patch token and the cls token for each head is extracted and denoted as $Att_h^{cls} \in {R^{N \times 1}}$. Corresponding total attention weights are calculated as follows:
\begin{equation}
A = \sum\limits_{h = 1}^H {Att_h^{cls}}
\end{equation}
The attention weight between the patch token in $(x,y)$ position and the cls token is denoted as ${A_{(x,y)}}$. For filtering out insignificant patches, the mean value ${\bar A}$ is calculated as the threshold and the new attention weight is as follows:

\begin{equation}
A_{(x,y)}^{new} = \left\{ {\begin{array}{*{20}{c}}
{{A_{(x,y)}}}&{\begin{array}{*{20}{c}}
{if}&{{A_{(x,y)}} > \bar A}
\end{array}}\\
0&{\begin{array}{*{20}{c}}
{otherwise}&{}
\end{array}}
\end{array}} \right.
\end{equation}

Inspired by ~\cite{zhou2020look}, polar coordinates are applied to measure the spatial relation between the most discriminative patch and other patch to mine object’s structure information. Specifically, the patch with the highest attention weight is regarded as the most discriminative patch, which is set as the reference patch. Given the reference patch ${P_0} = {P_{{x_0},{y_0}}}$ and the horizontal reference direction, where $({x_0},{y_0})$ are the indices in ${N_H} \times {N_W}$ plane, the polar coordinates of patch ${P_{x,y}}$ can be defined as:
\begin{equation}
{\rho _{x,y}} = \sqrt {{{(\frac{{x - {x_0}}}{{{N_W}}})}^2} + {{(\frac{{y - {y_0}}}{{{N_H}}})}^2}} 
\end{equation}
\begin{equation}
{\theta _{x,y}} = \frac{{(\arctan 2(y - {y_0},x - {x_0}) + \pi )}}{{2\pi }}
\end{equation}
where ${\rho _{x,y}}$ is the relative distance between ${P_{x,y}}$ and ${P_0 }$ and $\theta _{x,y}$ is the normalized polar angle of ${P_{x,y}}$ relative to the horizontal direction. To introduce this object structure information, we design the graph convolutional neural network to obtain the object structure feature. We firstly construct the graph which contains two components: (i) the image patch node features $X$ which depict the spatial context correlation based on the calculation of the polar coordinates and (ii) the edge weights obtained based on attention weight calculation between cls token and image patch token in the vision transformer layer, which summarize the significance of image patch token. Specifically, the matrix $Adj = A_{}^{new} \times {(A_{}^{new})^T}$ denotes the edge weights among nodes based on the ${A^{new}}$, where the edge weights related to insignificant patches are zeros to filter out their affect. The graph convolution is then adopted to further extract and incorporate the structure information into the vision transformer. The structure features $S$ are obtained by two-layer graph convolution as follows:
\begin{equation}
S = \sigma (Adj \times \sigma (Adj \times X \times {W^1}) \times {W^2})
\end{equation}
where ${W^1}$ and ${W^2}$ are learnable parameters and $\sigma ( \cdot )$ is activation function. The feature of the reference patch node is regarded as the object structure feature, which is further added to the cls token feature for introducing structure information into the transformer backbone. Through the end-to-end training, the composition of the object can be modeled and the significant image patch can be highlighted, which improves the model’s classification performance.

Overall, the structure information learning module can incorporate the object structure information, i.e., its spatial composition of key discriminative patches, into one or more vision transformer layers. The transformer network can be empowered to learn both appearance and structure information for accurate fine-grained classification.

\subsection{Multi-level Feature Boosting}\label{MFB}

After obtaining image features by the transformer feature extractor, the feature of the cls token $z_L^{cls}$ of the last transformer layer is generally selected as the image representation for final classification in vision transformer methods, such as ViT. However, it generally ignores the complementary of different levels’ features and the utilization of the inherent intra-class and inter-class semantic relations. Thus, we propose the multi-level feature boosting (MFB) module for enhancing feature robustness, which is shown in the left part of the Figure \ref{framework}.

Specifically, the features of cls tokens from the last three transformer layers are concatenated as the final image feature representation, which all introduce structure information through the structure learning module. In this simple but effective way, their complementary can be exploited to bring performance gains. To take full advantage of the semantic relations for feature enhancement, contrastive learning is adopted to enhance the feature similarity of the same category and weaken that of different categories. To mine the hard negative sample pairs to contribute to the model’s training, a hyper-parameter $\alpha $ is adopted in the following contrastive learning loss. The negative pairs with similarity which is $\alpha $ smaller than that of the positive pairs are filtered out. Formally, the contrastive learning loss on a batch of size $N$ is defined as follows:
\begin{equation}
\label{alpha}
\begin{array}{l}
Indicato{r_{i,j}} = \max \{ 0,\alpha  + sim({z_i},{z_j}^ - )\\
{\rm{                      }} - \frac{1}{{{\Gamma _{y(i) = y(j),i \ne j}}}}\sum\limits_{j:i \ne j} {sim({z_i},{z_j}^ + )} \} 
\end{array}
\end{equation}
\begin{equation}
\begin{array}{c}
{L_{CL}} = \frac{1}{{{N^2}}}\sum\limits_{i = 1}^N [ \sum\limits_{j:y(i) = y(j)}^N {(1 - sim({z_i},{z_j}^ + )} )\\
{\rm{         }} + \sum\limits_{j:y(i) \ne y(j)}^N {Indicato{r_{i,j}} \times sim({z_i},{z_j}^ - )} ]
\end{array}
\end{equation}
where ${({z_i},{z_j}^ + )}$ denotes the positive image representation pair which have the same category label, i.e., $y(i) = y(j)$, ${({z_i},{z_j}^ - )}$ denotes the negative image representation pair which belongs to different categories, ${{\Gamma _{{y_i} = {y_j},i \ne j}}}$ denotes the number of positive pairs, $sim( \cdot )$ denotes the cosine similarity calculation.

In summary, the vision transformer backbone, structure information learning module and feature boosting module of the proposed method are jointly trained end to end. The total objective is as follows:
\begin{equation}
L = {L_{CE}} + {L_{CL}}
\end{equation}

\section{Experiments} \label{exper}

In this section, we evaluate the performance of the proposed SIM-Trans approach on two standard fine-grained visual categorization datasets, i.e., the typical CUB-200-2011 \cite{cub200} dataset and the large-scale iNaturalist 2017 \cite{van2018inaturalist} dataset. We firstly introduce the datasets and evaluation metric. Then, we show specific implementation details, comparison experiments and analyses with recent state-of-the-art methods for each dataset respectively. Finally, in order to verify the effectiveness of the proposed modules, we conduct ablation experiments and analyses.

\subsection{Datasets and Evaluation Metric}
\label{Datasets and Evaluation Metric}

Two standard fine-grained visual categorization benchmarks are adopted in the experiments: 
\begin{itemize}[leftmargin=30pt]
\item {CUB-200-2011 \cite{cub200} is the most widely used dataset in the fine-grained visual categorization task. This dataset consists of 11788 images of 200 bird subcategories, where 5994 images are selected as training set and 5794 images are selected as testing set.} 
\item {iNaturalist 2017 \cite{van2018inaturalist} is one of the largest fine-grained visual categorization datasets. it contains more than 5,000 fine-grained categories and more than 95,000 test images. The detailed statistics are shown in table \ref{table_inat2017}, which includes category number and the split of training data and testing data. The dataset is very challenging for its large-scale fine-grained category number and testing data, which provides a strong benchmark for fine-grained visual categorization performance contrast.
}
\end{itemize}

Accuracy is adopted as the evaluation metric, which is generally used in fine-grained visual categorization task. And the definition is as follows:
\begin{equation}
Accuracy = \frac{{|{I_{{\rm{right}}}}|}}{{|I|}}
\end{equation}
where $|I|$ denotes the number of images in the testing set and $|{I_{{\rm{right}}}}|$ denotes the number of images that are correctly classified.

\begin{table}[!t]
 \centering
 \caption{Introduction of the iNaturalist 2017 dataset.}
 \label{table_inat2017}
 \begin{tabularx}{\linewidth}{|p{2cm}|Y|Y|Y|}
  \hline
  
  Super Class    & Class & Train Images & Test Images  \\ \hline
 
  Plantae    & 2101 & 158407 & 38206\\
  
  Insecta    & 1021 & 100479 & 18076\\

  Aves    & 964 & 214295 & 21226\\
  
  Reptilia    & 289 & 35201 & 5680\\

  Mammalia    & 186 & 29333 & 3490\\

  Fungi   & 121 & 5826 & 1780\\ 
  Amphibia    & 115 & 15318 & 2385\\ 
  Mollusca    & 93 & 7536 & 1841\\
  Animalia    & 77 & 5228 & 1362 \\
  Arachnida    & 56 & 4873 & 1086 \\
  Actinopterygii    & 53 & 1982 & 637\\
  Chromista    & 9 & 398 & 144\\
  Protozoa    &  4 & 308 & 73 \\ \hline
  Total    & 5089 & 579184 & 95986 \\
  
  \hline
  
 \end{tabularx}
\end{table}

\subsection{Experiments and Analyses on CUB-200-2011}
\label{ Experiments on the CUB-200-2011}

\subsubsection{\textbf{Implentation Details}}

In the experiments, the images are resized into $600 \times 600$ and randomly cropped into the size of $448 \times 448$ as input during the training stage. Random horizontal flip and normalization are used for preprocessing. All the above steps are standard pre-process steps which are generally used in fine-grained visual categorization methods, such as ~\cite{hu2021rams}. In the testing phase, the images are resized into the size of $600 \times 600$ and cropped into the size of $448 \times 448$ from the center. We adopt ViT as the backbone and the initial weights are loaded from the official ViT-B\_16 model pre-trained on the ImageNet 21k dataset, which is also adopted in RAMS-Trans~\cite{hu2021rams}. Thus, the comparison experiments with state-of-the-art (SOTA) transformer-based methods are fair and convincing. The image is split into patches of size 16 and the sliding window step is set as 12, which are $P$ and $S$ in Eq. \ref{slide} respectively. We adopt the stochastic gradient descent (SGD) optimizer with a momentum of 0.9 for model optimization. The learning rate is initialized as 3e-2 and cosine annealing schedule ~\cite{loshchilov2016sgdr} is exploited to update the learning rate. The batch size is set as 5 and the number of total training steps is set to be 10000 and the first 500 steps are warm-up. We adopt the last three transformer layers to plug in the structure information learning module in Sec. \ref{SIL} and the hyper-parameter $\alpha $ in Eq. \ref{alpha} is set to be 0.3. We perform the experiments with PyTorch using NVIDIA GeForce GTX 1080 Ti GPUs.

\begin{table}[!t]
 \centering
 \caption{Comparison experiments with other state-of-the-art methods on CUB-200-2011 dataset.}
 \label{table_cub_acc}
 \begin{tabularx}{\linewidth}{|p{4.3cm}|Y|Y|}
  \hline
  
  Method    & Backbone & Acc(\%)  \\ \hline

  KP~(CVPR 2017)~\cite{cui2017kernel}    & VGG16 &86.2 \\
  MA-CNN~(ICCV 2017)~\cite{zheng2017learning}    & VGG19 &86.5 \\
  PC~(ECCV 2018)~\cite{dubey2018pairwise}   & DenseNet161 &86.9 \\
  NTS-Net~(ECCV 2018)~\cite{yang2018learning}    & ResNet50 &87.5 \\

  M2DRL~(IJCV 2019)~\cite{he2019and}    & VGG16 &87.2 \\

  S3N~(ICCV 2019)~\cite{ding2019selective}    & ResNet50 &88.5 \\
  FDL~(AAAI 2020)~\cite{liu2020filtration}    & ResNet50 &88.6 \\
  LIO~(CVPR 2020)~\cite{zhou2020look}    & ResNet50 &88.0 \\
  
  PMG~(ECCV 2020)~\cite{du2020fine}    & ResNet50 &89.6 \\   
  
  DP-Net~(AAAI 2021)~\cite{wang2021dynamic}  & ResNet50 &89.3 \\

  GaRD~(CVPR 2021)~\cite{zhao2021graph}    & ResNet50 &89.6 \\ 
  Chang et al.~(CVPR 2021)~\cite{chang2021your}    & ResNet50 &89.9 \\ 
  SPS~(ICCV 2021)~\cite{huang2021stochastic}    & ResNet50 &88.7 \\ 
  Joung et al.~(ICCV 2021)~\cite{joung2021learning}    & ResNet50 &88.4 \\ 
  CAL~(ICCV 2021)~\cite{rao2021counterfactual}    & ResNet101 &90.6 \\
  MCEN~(ACM MM 2021)~\cite{li2021multi}    & ResNet50 &89.3\\ 
  ViT~(ICLR 2020)~\cite{dosovitskiy2020image}    & ViT-B\_16 &90.6\\ 

  RAMS-Trans~(ACM MM 2021)~\cite{hu2021rams}    & ViT-B\_16 &\underline{91.3}\\
  \textbf{Our SIM-Trans approach}   & \textbf{ViT-B\_16} & \textbf{91.8} \\
  
  \hline
  
 \end{tabularx}
\end{table}

\subsubsection{ \textbf{Comparisons with State-of-the-art Methods }}

This subsection presents comparison experimental results and analyses with other state-of-the-art (SOTA) methods including CNN based methods and transformer based methods on CUB-200-2011. For fair comparison, all the methods adopt the same training and testing setting, such as the same input image resolution. From Table \ref{table_cub_acc}, we can observe that:

\begin{itemize}[leftmargin=30pt]
\item {On CUB-200-2011, our proposed SIM-Trans approach achieves the best classification accuracy of 91.8\%, which brings 1.2\% improvement than the ViT baseline method. Compared with the optimal CNN based method CAL~\cite{rao2021counterfactual}, our SIM-Trans achieves 1.2\% performance improvement. In the CNN based methods, attention mechanism and multi-scale learning are widely used for discriminative information mining and feature robustness enhancement. For example, the counterfactual attention learning method is proposed to learn more effective attention for fine-grained classification in CAL. In PMG \cite{du2020fine}, mutli-granularity information with jigsaw operation is introduced to boost model’s recognizing from different scales. However, the above CNN based methods are generally affected by the spatial resolution degradation, which is caused by the convolutional strides and pooling operation. The transformer based methods can keep the patch information with the deeper of the layers, which can keep discriminative detailed information. Thus, the transformer based method have the advantage for fine-grained visual categorization and even the pure vision transformer ViT baseline can perform as well as the SOTA CNN based method, achieving 90.6\% classification accuracy.
} 

\item {Compared with the transformer based methods, our SIM-Trans also show the best performance, outperforming the RAMS-Trans by a margin of 0.5\%. RAMS-Trans proposes to use the attention weights to guide the model to amplify discriminative regions for multi-scale learning, which is a heuristic work to combine the advantages from both the CNN and transformer. However, the multi-branch framework increases the computation cost and the parameter-sharing in different scales may cause the model’s confusion. By contrast, our SIM-Trans guides the model to focus on the holistic object structure and its component discriminative patches. It is a soft way to boost the significant patches and inhibit the useless patches, such as the background patch. Our SIM-Trans approach makes an attempt to empower the model to conduct fine-grained recognition from both the object structure and the local discriminative patches. Learning both the object structure information and fine-grained appearance information makes our SIM-Trans approach achieve better classification performance with strong fine-grained representation ability.}

\begin{figure}[!t]
  \centering
  \includegraphics[width=1.0\linewidth]{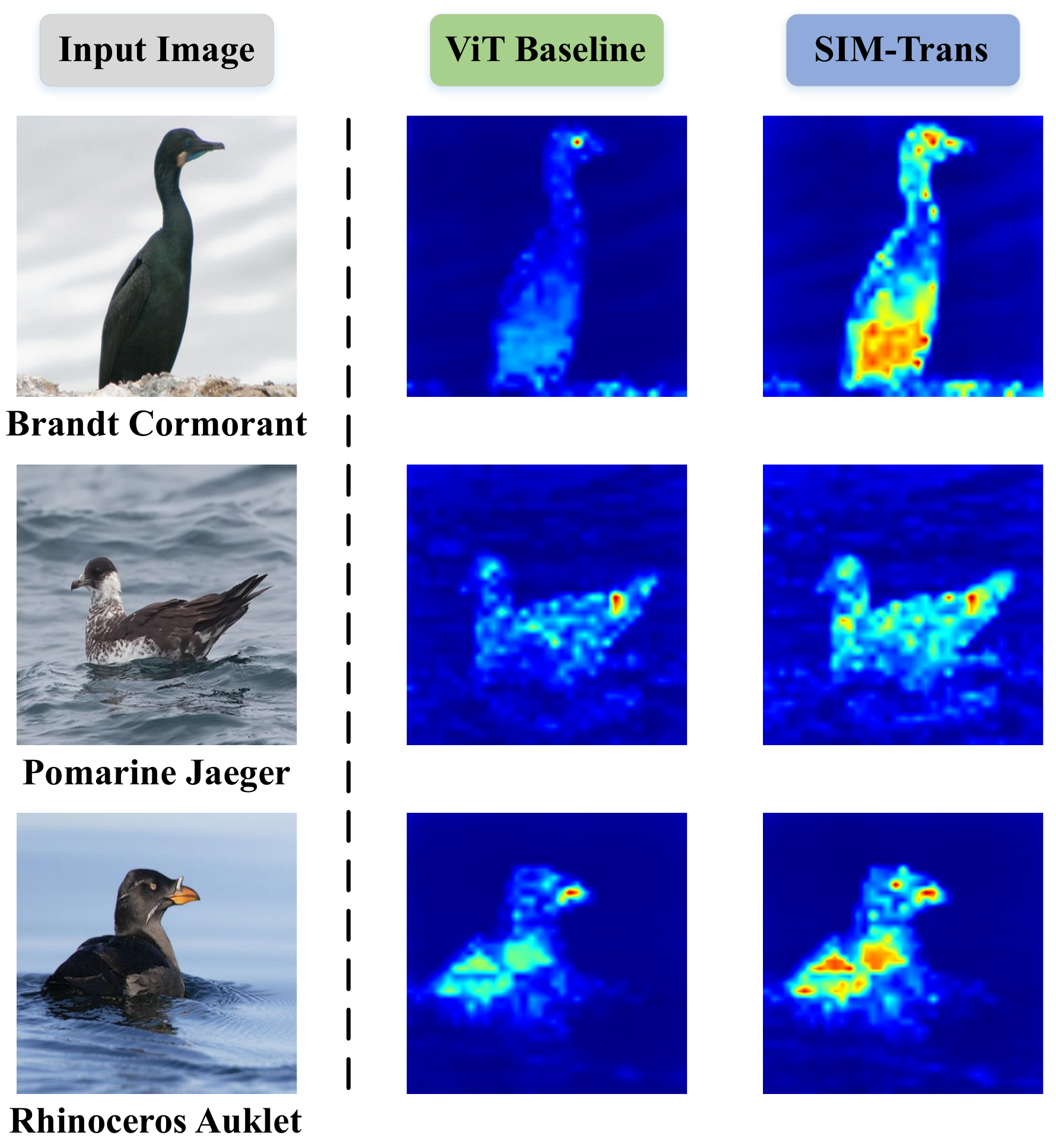}
  \caption{Attention visualization based on ViT baseline and our proposed SIM-Trans. The first column is original images and the second column shows the attention generated by ViT baseline. The last column shows the attention generated by SIM-Trans.}
  \label{motivation}
\end{figure}

\end{itemize}

\begin{figure*}[!t]
  \centering
  \includegraphics[width=\linewidth]{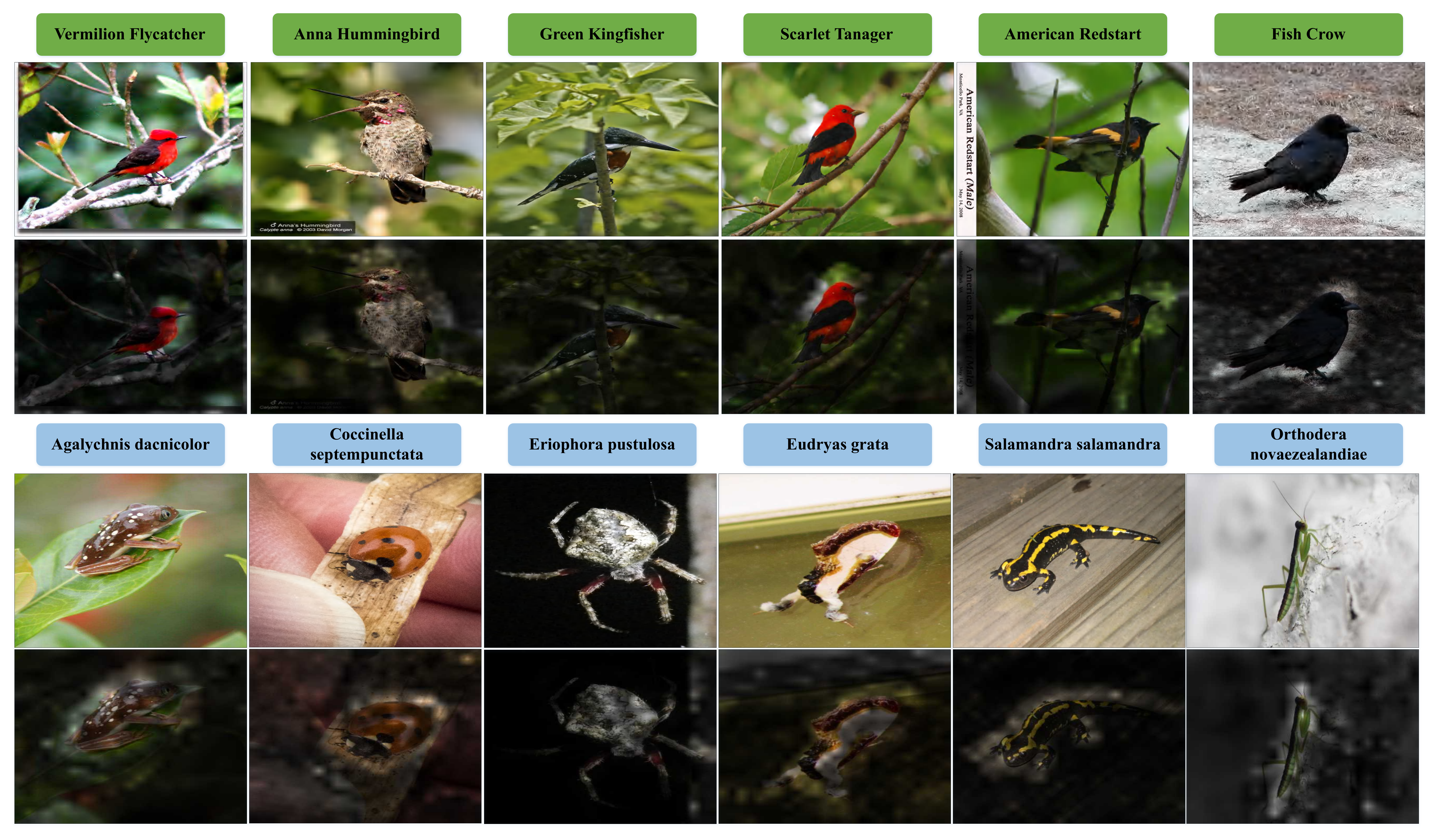}
  \caption{Visualization results of SIM-Trans on CUB-200-2011 and iNaturalist 2017 datasets. The first row and the third row are original images while the second and the fourth rows present focus regions generated by our SIM-Trans model.}
  \label{global_attention}
\end{figure*}

\subsubsection{\textbf{Qualitative Results }}

As shown in Figure \ref{motivation}, compared with the ViT baseline, our proposed SIM-Trans can not only filter out irrelevant background information but also complete the object extent and highlight discriminative regions, which verifies its effectiveness in discriminative information mining.

In order to visualize the structure learning effect of our SIM-Trans approach, we present the raw images and focus regions of our SIM-Trans model in Figure \ref{global_attention}. The first and third rows show the raw images sampled from the CUB-200-2011 dataset and iNaturalist 2017 dataset respectively. And the second and fourth rows are corresponding focus regions. From Figure \ref{global_attention}, we can observe that the entire object and discriminative regions are highlighted precisely, which verifies the effectiveness and improve the interpretability of our proposed SIM-Trans approach.

\subsection{Experiments and Analyses on iNaturalist 2017}

\begin{table}[t]
 \centering
 \caption{Comparison experiments with other state-of-the-art methods on iNaturalist 2017 dataset.}
 \label{table_inat_acc}
 \begin{tabularx}{\linewidth}{|p{4.5cm}|Y|Y|}
  \hline
  
  Method    & Backbone & Acc(\%)  \\ \hline
 
  ResNet152~(CVPR 2016)~\cite{he2016deep}    & ResNet152 &59.0\\
  
  IncResNetV2~(AAAI  2017)~\cite{szegedy2017inception}    & InResNetV2 &67.3\\

  SSN~(ECCV 2018)~\cite{recasens2018learning}    & ResNet101 &65.2\\
  
  TASN~(CVPR 2019)~\cite{zheng2019looking}    & ResNet101 &68.2\\

  Huang et al.~(CVPR 2020)~\cite{huang2020interpretable}    & ResNet101 &66.8\\

  ViT~(ICLR 2020)~\cite{dosovitskiy2020image}    & ViT-B\_16 &67.0\\ 
  RAMS-Trans~(ACM MM 2021)~\cite{hu2021rams}    & ViT-B\_16 &\underline{68.5}\\
  \textbf{Our SIM-Trans approach}  & \textbf{ViT-B\_16} & \textbf{69.9} \\
  
  \hline
  
 \end{tabularx}
\end{table}

\subsubsection{\textbf{Implementation Details}}

For fair comparison with other methods, the size of images is set as $304 \times 304$ following ~\cite{hu2021rams}. We load the weights from the official ViT-B\_16 model pre-trained on the ImageNet 21k dataset for fair comparisons with other SOTA transformer based methods. The batch size is set to be 16 and the number of total training steps is set to be 100000. Stochastic gradient descent (SGD) optimizer with a momentum of 0.9 is adopted, and the learning rate is set as 1e-2 with cosine annealing scheduler. We introduce the structure information in the last three transformer layers and the hyper-parameter $\alpha $ in Eq. \ref{alpha} is set as 0.3.

\subsubsection{ \textbf{Comparisons with State-of-the-art Methods }}

To fully validate the effectiveness of our proposed SIM-Trans approach for fine-grained visual categorization, comparison experiments with SOTA methods on the large-scale fine-grained visual categorization benchmark iNaturalist 2017 are conducted. Table \ref{table_inat_acc} summarizes the result, and we can observe that:
\begin{itemize}[leftmargin=30pt]
\item {On iNaturalist 2017, our proposed SIM-Trans approach achieves the best performance with 69.9\% classification accuracy, which outperforms the pure resnet152 and the pure ViT (our baseline) by 10.9\% and 2.9\% respectively. Compared with the optimal CNN based method TASN and transformer based method RAMS-Trans, our SIM-Trans achieves 1.7\% and 1.4\% improvements respectively. Thanks to the simplicity and effectiveness of introducing structure information into vision transformer, our SIM-Trans approach is able to extend to both academic dataset and large-scale challenging scenario. The above comparison experiment results verify the effectiveness of our proposed SIM-Trans approach on the large-scale FGVC benchmark, whose categories are more than 5,000 with more than 95,000 test images.
} 

\end{itemize}

\subsubsection{\textbf{Qualitative Results }}

The visualization results of our SIM-Trans approach on iNaturalist 2017 are also shown in Figure \ref{global_attention}. Visual objects of different subcategories belonging to different super categories can be highlighted precisely, which verifies the accurate object structure modeling and generalization ability of our proposed SIM-Trans model.

\begin{table}[th]
 \centering
 \caption{Ablation Experiments on CUB-200-2011 dataset.}
 \label{table_ablation}
 \begin{tabularx}{\linewidth}{|p{5cm}|Y|}
  \hline
  
  Method   & Acc(\%)  \\ \hline
 
  Baseline   &90.6\\
  
  Baseline + SIL    &91.1\\  
  
  Baseline + SIL + MFB\_without\_CL   &91.4\\

  \textbf{Baseline + SIL + MFB} &\textbf{91.8}\\

  \hline
  
 \end{tabularx}
\end{table}

\subsection{Effectiveness of Each Component in Our SIM-Trans Approach}

\begin{figure}[!t]
  \centering
  \includegraphics[width=\linewidth]{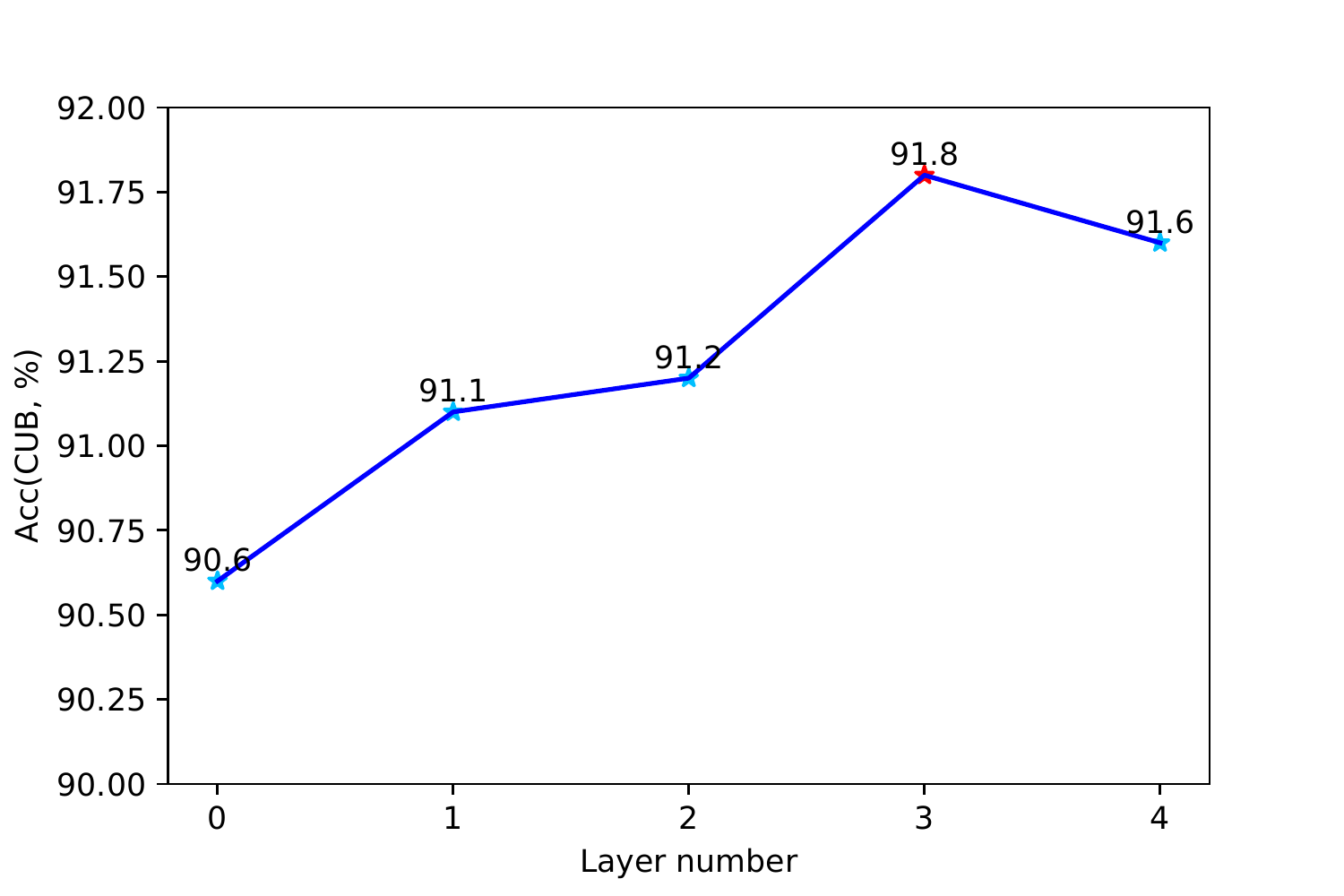}
  \caption{Layer number experiments of structure information introduction on CUB-200-2011 dataset.}
  \label{layer}
\end{figure}

Structure information learning (SIL) module and multi-level feature boosting (MFB) module are proposed in our SIM-Trans approach. We conduct ablation experiments on CUB-200-2011 dataset and the effectiveness of each component is shown in Table \ref{table_ablation} and Figure \ref{layer}, we can observe that:

\begin{itemize}[leftmargin=30pt]
\item {A classifier head is added to the ViT extractor as our baseline, which can achieve 90.6\% classification accuracy on CUB-200-2011 dataset. Compared with our baseline, the structure information learning (SIL) module improves the classification accuracy by a margin of 0.5\%, which is simply added to the last transformer layer as a plug-in. The SIL module boosts the model’s understanding ability for the object structure and highlights significant regions to make the feature representation more discriminative.
}

\item {Based on the SIL module, we conduct multi-layer structure information injection and the multi-level feature boosting (MFB) for robust feature learning, which can bring another 0.7\% performance gain. We attempt to add the SIL module from the top layer to the bottom layer, and Figure \ref{layer} shows when the number of layers is set as three, the proposed model can achieve the best performance. It also demonstrates the combination of SIL’s expansion and MFB’s fusion contributes to each other for improving the classification performance.
}
\item {In MFB, the contrastive learning can improve the classification accuracy by a margin of 0.4\%, which attends to hard negative samples and utilize the semantic relations to boost the model's feature discrimination.}

\end{itemize}

\subsection{Hyper-parameter Experiments}

\begin{figure}[t]
  \centering
  \includegraphics[width=\linewidth]{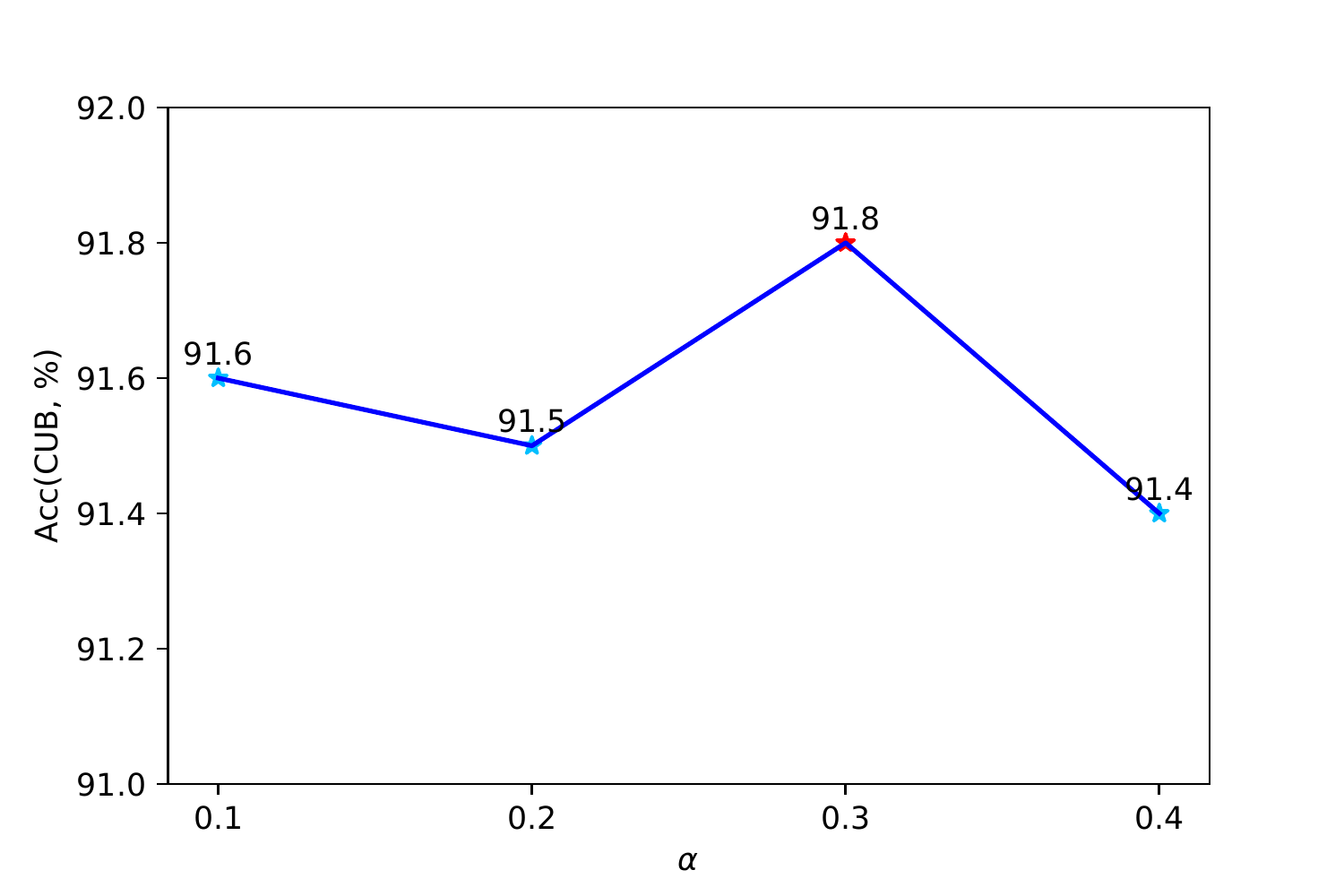}
  \caption{Hyper-parameter experiments of $\alpha $ on CUB-200-2011 dataset.}
  \label{canshu}
\end{figure}

There is only one hyper-parameter in our SIM-Trans model designing. The $\alpha $ of MFB module in Eq. \ref{alpha} controls the hardness degree of negative pairs in contrastive learning. We analyze the influence of $\alpha $ on the classification performance in Figure \ref{canshu}. When $\alpha $ is set to be 0.3, the proposed SIM-Trans model can achieve the best classification performance. Continuing to increase $\alpha $, the performance degrades because the over-high threshold cannot filter out easy negative pairs, which is harmful for contrastive learning.

\section{Conclusion}
In this paper, we propose the structure information modeling transformer (SIM-Trans) that introduces the object structure information into vision transformer for boosting the discriminative feature learning to contain both the appearance and structure information. Structure information learning (SIL) module is proposed to mine spatial context relation of significant patches within the object extent, which boosts model’s understanding ability for object structure and highlights discriminative regions. Multi-level feature boosting (MFB) module is then proposed to exploit the complementary of multi-level features and contrastive learning to further enhance feature representation robustness for accurate fine-grained recognition. The combination of the proposed two modules boosts each other and promotes the feature discrimination. The proposed SIM-Trans approach provides an attempt to model object structure in transformer framework and achieves state-of-the-art for FGVC task on two typical fine-grained benchmarks, especially achieving promising performance on the large-scale iNaturalist 2017 benchmark.

In the future, the structure embedding to the initial layer of the transformer framework will be studied for better structure modeling to improve fine-grained visual categorization performance.

\section{Acknowledgments}
This work was supported by the grants from the National Natural Science Foundation of China (61925201, 62132001, U21B2025) and the National Key R\&D Program of China (2021YFF0901502).

\newpage
\balance

\bibliographystyle{ACM-Reference-Format}

\bibliography{sample-base}


\end{document}